\documentclass[11pt]{article} 

\usepackage{times}  
\usepackage{helvet} 
\usepackage{courier}  
\usepackage[hyphens]{url} 
\usepackage{graphicx}
\usepackage{bm}
\usepackage{caption} 
\usepackage{subfig}
\usepackage{a4wide}
\usepackage{algorithm}
\usepackage{algorithmic}
\usepackage{mathtools} 
\usepackage{newfloat}
\usepackage{listings}
\DeclareCaptionStyle{ruled}{labelfont=normalfont,labelsep=colon,strut=off} 
\floatstyle{ruled}
\newfloat{listing}{tb}{lst}{}
\floatname{listing}{Listing}

\usepackage[width=6.5in, height=9.65in, head=0.0in, foot=0.3in, headsep=0.0in]{geometry}
\usepackage{authblk}
\usepackage{physics}
\usepackage{amsmath}

\title{Improving Knowledge Distillation in Transfer Learning with Layer-wise Learning Rates}

\date{}
\author{\normalsize\textbf{Shirley Kokane}\textsuperscript{1}, \normalsize\textbf{Mostofa Rafid Uddin}\textsuperscript{1}, \textbf{Min Xu}\textsuperscript{1,$\dagger$} \\\textsuperscript{1} 
Carnegie Mellon University, Pittsburgh, PA 15213, USA \\ 
$\dagger$ \emph{} Corresponding Author: \texttt{mxu1@cs.cmu.edu}}

\usepackage{bibentry}

\begin{document}

\maketitle

\begin{abstract}
Transfer learning methods start performing poorly when the complexity of the learning task is increased. Most of these methods calculate the cumulative differences of all the matched features and then use them to back-propagate that loss through all the layers. Contrary to these methods, in this work, we propose a novel layer-wise learning scheme that adjusts learning parameters per layer as a function of the differences in the Jacobian/Attention/Hessian of the output activations w.r.t. the network parameters. We applied this novel scheme for attention map-based and derivative-based (first and second order) transfer learning methods. We received improved learning performance and stability against a wide range of datasets. From extensive experimental evaluation, we observed that the performance boost achieved by our method becomes more significant with the increasing difficulty of the learning task.
\end{abstract}

\section{Introduction}
\label{sec:Introduction}
Neural networks have achieved massive success in fields like computer vision and natural language processing \cite{paper_24, paper_25, paper_26}. There is continual progress being made with extensive state-of-the-art performances. Despite the massive success, deep neural networks face two major disadvantages in real-world applications. Firstly, deeper neural networks are highly compute-intensive and difficult to deploy in resource-limited situations. Secondly, it requires extensive data to make such large networks achieve higher task accuracy, which is unavailable in most scenarios.

Knowledge transfer learning methodologies have received a surge in popularity among the machine learning community to tackle the above-mentioned challenges. These methodologies enable training one neural network, often referred to as the student model, with the help of another neural network, often referred to as the teacher model, to train on the same dataset or a subset of a similar dataset. Knowledge distillation is the type of knowledge transfer method where the teacher network is larger than the student network and both networks are trained on the same dataset \cite{paper_17, paper_18}. Knowledge distillation is usually applied in machine learning to extract the most important aspects of the teacher neural network model into a less computation-intensive student model. It involves the concept of passing on knowledge gained by important neurons in larger (teacher) networks to the smaller (student) networks so that the smaller networks learn crucial features contributing to achieving better performance \cite{paper_4}. 

Recent knowledge distillation approaches focus on utilizing derivative information of neural network parameters to obtain the best approximation of the model \cite{paper_7,paper_8,paper_9}. On the other hand, there have been several other attention-matching methods that match the spatial maps obtained post each critical neural network activation \cite{paper_2,paper_3}. However, the existing methods of these types utilize the cumulative difference of all the maps and then use it to back-propagate that loss through all the layers. The cumulative loss back-propagation is often ineffective and results in student models that do not achieve the desired accuracy compared to their respective teacher models. This problem becomes more intense with the increased complexity of the learning task.

\begin{figure*}[t]
    \centering
    \includegraphics[height=0.5\linewidth,width=1.0\linewidth]{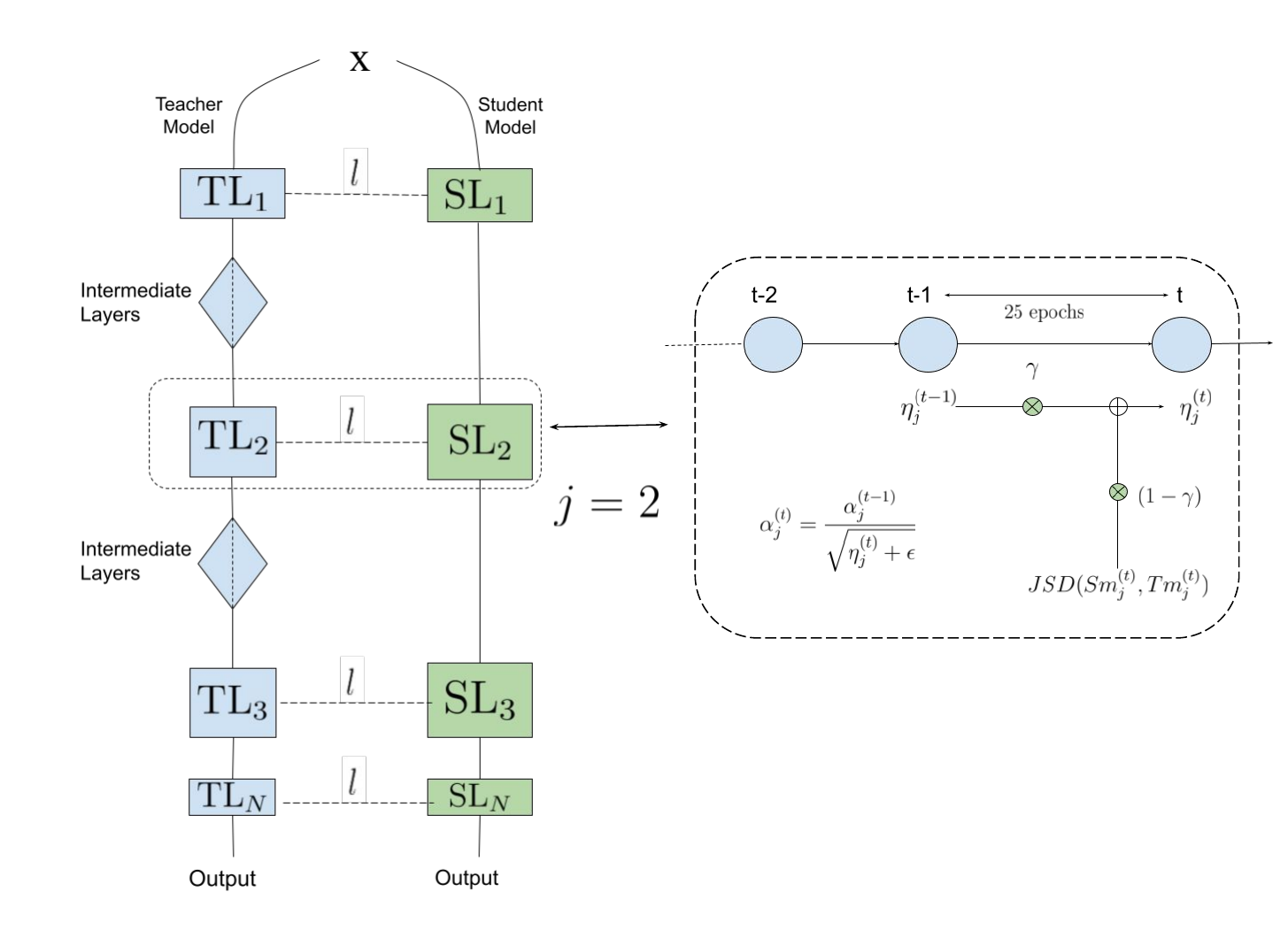}
    \caption{Schematic representation of the layer-wise learning rate optimization. On the left, $TL_N$ represents teacher layers and $SL_N$ represents the corresponding layers in the student model that dimensionally match with the teacher; the corresponding loss between them is represented by $\it{l}$. The teacher model has more layers than the student model. On the right are the steps of updating the learning rate ($\alpha_j^{(t-1)}$) based on the JSD loss (between the student and the teacher layer at the second identical layer) and prior momentum ($\eta_j^{(t-1)}$) at an interval of 25 epochs.}
    \label{fig:method}
\end{figure*}

In this work, we have dealt with these problems by implementing layer-wise learning rates with respect to the differences in the network weight derivates or attention maps. We first select the crucial layers that dimensionally match the student and the teacher model. Subsequently, we compute the spatial map via attention or order-based mapping to compare the student and teacher models. Then we calculate the loss between these layer-wise maps to attain a corresponding loss of each crucial layer. Finally, we utilize these per-layer losses to tune the learning rate of these crucial layers depending on the magnitude of the loss. We present a layer-specific training approach for transfer learning that is investigated over attention mapping, first-order mapping, and second-order mapping.

Moreover, earlier research has focused more on taking approximations (first/second order) mainly with respect to the inputs, which is not representative of the changes made in the architecture of the network \cite{paper_3}. In this work, we focus on comparing the first and second-order derivatives of the network parameters with respect to the weights of the neural network layers.

We experimented with our method on CIFAR and CoCo benchmark datasets. Our experimental results demonstrate that using individual layer-specific learning rates effectively improves the model's performance, leading to better convergence. We also observe that the order-based mapping that is Jacobian matching performs significantly better when the task complexity increases. We propose better-converging methods that work effectively on harder tasks and profoundly generalize over test data.

Our main contributions are as follows: 

\begin{itemize}
    \item We studied on the effects of Layer-wise Learning on Attention Matching Transfer Learning, Jacobian Matching Transfer Learning, and Hessian Matching Transfer Learning on publicly available datasets like CIFAR10, CIFAR100, and CoCo.
    
    \item We have identified the critical layers essential for the loss and layer-specific learning rate calculations.
    
    \item We have further analyzed and demonstrated state-of-the-art performances caused by applying layer-specific learning rates for each of the attention-based, jacobian-based, and Hessian-based methods.
    
\end{itemize}

\section{Related Work}
\label{sec:related_work}

A common method in transfer learning involves training a large-scale teacher model on a broad task and then leveraging the learned information to train a smaller student model \cite{paper_5}. This technique aligns with the Learning without Forgetting principle, where a smaller model is trained to adapt to tasks while trying to match the output responses of the original (teacher) model \cite{paper_6}. By compressing information into smaller networks, we can eliminate redundant neurons, enhancing the efficiency of network pruning \cite{paper_10,paper_11,paper_12}. In the following, we demonstrate several prominent transfer learning methods, that have gained major recognition in this domain:

\paragraph{\textbf{Attention based methods}:  }

Attention mechanisms have been described diversely across various studies.
Zagoruyko et. al., differentiate between activation-based and gradient-based attention maps, aiming to rank these maps based on the quality of information they capture from images \cite{paper_2}. Similarly, Wang et. al., have developed a method to assign weights to output activation maps, employing feature boosting and suppression to manipulate the loss effectively  \cite{paper_4}. 

\paragraph{\textbf{Derivative based methods}: }
Several works have leveraged order-based information for different problems, including neural network approximation. This is particularly common in reinforcement learning, where training aims to align the critic’s derivatives with target derivatives \cite{paper_13}. In this paper, we particularly focus on the advantages of derivative maps in aligning student models with teacher models by matching their derivative structures, thereby enhancing knowledge transfer.  \\
\newline
There are two types of derivative-based methods, we discuss them as follows:

\paragraph{\textbf{Jacobian based methods}:}
The use of Jacobians in model distillation has been studied for its potential to enhance performance. Czarnecki et al. utilized the derivative of each layer’s output concerning the input, suggesting approximations via projection over a random vector to address computational intensity \cite{paper_3}.  The concept of matching gradient-based attention maps was initially explored by Zagoruyko \& Komodakis (2017), who considered the Jacobian of inputs with respect to outputs as one such attention map. The Jacobian of inputs with respect to the outputs was also considered one such attention map. They found that combining activation-based attention maps with Jacobian maps improved performance. Jacobian-norm regularizers, which penalize the Jacobian norm to make models more robust to small input changes, were introduced by \cite{paper_16}. 

\paragraph{\textbf{Hessian based methods}:}

Although crucial in understanding neural network convergence, second-order derivative information has not been widely explored in transfer learning. It is extensively used to predict generalization rates, enhance generalization through weight elimination, and in full second-order optimization methods by \cite{paper_22, paper_23}.  
Zhang et al. demonstrated the advantages of using local Hessians for backpropagation over first-order gradients. They showed that second-order approximations help in training deep networks and improve performance when combined with techniques like skip connections and batch normalization. They also developed a new stochastic gradient descent algorithm utilizing the local Hessian of the network \cite{paper_14}. Wu et al. highlighted that the top eigenspace of Hessians is significantly structured, showing substantial coincidence across models trained with different initializations, and that top eigenvectors resemble rank-1 matrices when reshaped to match the corresponding weight matrix \cite{paper_15}. 

All the above-described methods offer broad optimization by using overall derivative maps calculated based on the raw inputs. This does not sufficiently enhance the individual layer-wise learning required for each layer to align uniformly with its teacher layer. We address this issue by introducing individual layer-wise losses to bridge the gap between each student layer and its corresponding teacher layer.

\section{Method}

Considering a neural network model $\boldsymbol{m}$ parameterized with $\boldsymbol{\theta}$, one typically seeks to minimize the empirical error between the proposed model outputs and the target $\boldsymbol{y}$ according to some loss function, 

\begin{equation}
\sum_{i=1}^{N} l_1\Bigr(m(x_i| \theta) , y_i \Bigr)
\end{equation}

where $\boldsymbol{N}$ is the number of training samples, $\boldsymbol{x_i}$ is the input for the ith training sample, and $\boldsymbol{l_1}$ is a loss function defined to quantify the disparity between target and prediction (eg. cross-entropy for classification problems)

\subsection{Layerwise learning rate for Attention map-based methods}

Attention maps have gained different approaches in the transfer learning world for better training of the student model. We intend to provide as much information to the student model from the teacher with fewer layers compared to the latter. Let us consider a student model $\boldsymbol{Sm}$ with $\boldsymbol{k}$ number of crucial layers that dimensionally match with the layers in the teacher model $\boldsymbol{Tm}$. We then calculate the total loss of these attention maps over all the training samples in the following way: 

\begin{equation}
\sum_{i=1}^{N} \Bigr[l_1(Sm(x_i) , y_i) + \sum_{j=1}^{k} l_2(Sm_j(x_i) , Tm_j(x_i))\Bigr]
\end{equation}

The $\boldsymbol{l_2}$ loss is a function best used to represent the difference in 2 distributions. We have proposed the use of Jensen–Shannon divergence in the later section. 

To update the learning rate of these crucial layers over a set of epochs, we define a learning rate optimizer function $\boldsymbol{g}$: 

\begin{equation}
    \alpha_j =  \it{g}\Bigr(\sum_{i=1}^{N} l_2(Sm_j(x_i) , Tm_j(x_i))\Bigr)
\end{equation}

The function $\boldsymbol{g}$ is a set of steps applied to the layer-wise loss which will be explained in the later section. 

\subsection{Layerwise learning rate for Jacobian map-based methods}

In Jacobian Matching with respect to weights, we calculate and minimize the loss function in the following way:

\begin{equation}
    \sum_{i=1}^{N} \Bigr[l_1(Sm(x_i) , y_i) + \sum_{j=1}^{k} l_2\Bigr(\dv {Sm_j(\theta)}{Sm(x_i)}, \dv {Tm_j(\theta)}{ Tm(x_i)}\Bigr)\Bigr]
\end{equation}

where $\dv {Sm_j(\theta)}{Sm(x_i)}$ is the derivative of the parameters of the j-th layers of the network with respect to the network output for the input ${x_i}$. 

Subsequently, we update the learning rate in equal intervals of the number of epochs, 

\begin{equation}
    \alpha_j =  \it{g}\Bigr(\sum_{i=1}^{N} l_2\Bigr(\dv {Sm_j(\theta)}{Sm(x_i)}, \dv {Tm_j(\theta)}{ Tm(x_i)}\Bigr)\Bigr)
\end{equation}

\subsection{Layerwise learning rate for Hessian map-based methods}

We observed several advantages of the second-order approximations over the Jacobian. Hence, we would be applying local Hessian as a knowledge transfer tool for more effective performance. 

Corresponding to the procedure we utilized in Jacobian Matching, we would be calculating the second-order derivative of weights for the cumulative loss in the following way:

\begin{equation}
    \sum_{i=1}^{N} \Bigr[l_1(Sm(x_i) , y_i) + \sum_{j=1}^{k} l_2\Bigr(\dv [2]{Sm_j(\theta)}{Sm(x_i)}, \dv [2]{Tm_j(\theta)}{ Tm(x_i)}\Bigr) \Bigr]
\end{equation}

where $\dv [2]{Sm_j(\theta)}{Sm(x_i)}$ is the second order derivative of the parameters of the j-th layers of the network with respect to the network output for the input ${x_i}$. 

Subsequently, we update the learning rate in equal intervals of the number of epochs, 

\begin{equation}
    \alpha_j =  \it{g}\Bigr(\sum_{i=1}^{N} l_2\Bigr(\dv [2]{Sm_j(\theta)}{Sm(x_i)}, \dv [2]{Tm_j(\theta)}{ Tm(x_i)}\Bigr)\Bigr)
\end{equation}

\subsection{Layerwise Learning Rate Optimization Function}

Our goal with transfer learning using attention and orderwise maps is to minimize the divergence between the distributions of these two maps at every layer. To precisely approximate the variance in these two distributions, we employ the Jensen-Shannon Divergence (JSD). 

Let $Sm_j^{(1)}$ and $Tm_j^{(1)}$ be the attention map or orderwise map of the jth shortlisted layer of the student and teacher model respectively for the $t_th$ iteration. Then the JSD between $Sm_j^{(1)}$ and $Tm_j^{(1)}$ is

\begin{equation}
    JSD(Sm_j^{(t)},Tm_j^{(t)})  = H(p) - \beta_1H(Sm_j^{(t)}) - \beta_2H(Tm_j^{(t)}) 
\end{equation}

\begin{equation}
\begin{split}
    JSD(Sm_j^{(t)},Tm_j^{(t)})  = \beta_1KLD(Sm_j^{(t)} || p) + \\ \beta_2KLD(Tm_j^{(t)} || p) 
\end{split}
\end{equation}

\begin{equation}
    p = {\beta_1}{Sm_j^{(t)}} + {\beta_2}{Tm_j^{(t)}}
\end{equation}

where \textit{H} is the Shannon Entropy and $\beta_1$ , $\beta_2$ are the weights given to each distribution. In this case, they would be equally weighted. \textit{KLD} is the KL-Divergence of the 2 distributions. 

The advantage of using JSD over MSE or KLD loss is that JSD is symmetric, well-bounded, and does not imply absolute continuity in the distributions \cite{paper_20,paper_19, paper_21}. We apply this JSD loss to update the learning parameter for each crucial layer individually and sequentially. We define $\boldsymbol{\alpha_j}$ as the learning rate for each crucial layer applied to the optimizer and $\boldsymbol{\eta_j}$ as the momentum to regularize the learning rate per layer. 

Here we have made the updates at the interval of every 25 epochs, thus the difference between $\boldsymbol{t}$ and $\boldsymbol{t-1}$ is 25 epochs. The momentum term monitors the updates made to the learning rate and avoids $\boldsymbol{\alpha}$ diminishing to decay rapidly due to larger loss values. The $\boldsymbol{\eta_j^{(t)}}$ values will refrain the updates in $\boldsymbol{\alpha_j^{(t)}}$ to have larger denominators, hence avoiding slower convergence. 

\begin{equation}
    \alpha_j^{(t)} = \frac{\alpha_j^{(t-1)}}{\sqrt{\eta_j^{(t)} + \epsilon}}
\end{equation}

\begin{equation}
    \eta_j^{(t)} = \gamma \eta_j^{(t-1)} + (1-\gamma)JSD(Sm_j^{(t)},Tm_j^{(t)})
\end{equation}

where $\epsilon$ is a constant provided to avoid dividing by an absolute zero and $\gamma$ is a constant between 0 - 1, deciding the weight given to the loss. 

\section{Experiments}

To demonstrate the efficacy of our method, we experimented with our proposed approaches against two CIFAR (CIFAR-10 and CIFAR-100) image classification datasets and CoCo multi-class object classification dataset \cite{cifar, paper_27}. Against each of the datasets, we performed attention map-based, Jacobian map-based, and Hessian map-based knowledge distillation with and without our proposed layerwise learning approach. 
We used Pytorch to create the deep-learning models and run the networks on NVIDIA RTX A5000 and AMD Radeon GPU machines. 

\subsection{CIFAR Experiments}

For the CIFAR experiments, we used ResNet18 as the teacher model and ResNet10 as the student model. In the CIFAR10 and CIFAR100 datasets, we have 10k images in the test set and 50k images in the train set. To make the Attention, Jacobian, and Hessian calculations more effective, we only calculate the maps of the crucial conversion layers of the ResNet where we observe a dimensional change in the feature channels. In this case, we have 4 such important layers. We use the CIFAR dataset Images for training both teacher and student model for 200 epochs. In the experiments involving the learning rate scheduler, we use the MultiStep LR at a decaying factor 0f 0.01 at the 25th and 35th epochs. This learning rate scheduler is used to compare the learning rates of all the model layers. 

For the layerwise learning rate experiments, we individualize the important layers in the optimizer and subsequently update the learning rate of the prominent layers at every 25th epoch.
On each of these four layers, we calculate attention, Jacobian, and Hessian maps. We intend to minimize the difference in these corresponding maps between the student and teacher models. Besides including these comparative losses with our initial cross-entropy loss, we have employed our proposed layer-wise loss to optimize the learning rate of these individual layers. 

We perform experiments across attention, jacobian, and hessian knowledge distillation methods with:
\begin{itemize}
    \item With constant Learning Rate. 
    \item Learning Rate Scheduler at constant interval (25th and 35th epoch). 
    \item Layerwise Learning Rate with Learning Rate Scheduler. 
\end{itemize}

\begin{table*}[h]
    \centering
    \begin{tabular}{|c|c|c|c|}
    \hline
        Method &  Learning Rate update & CIFAR10 Accuracy & CIFAR 100 Accuracy\\
        \hline
         & None &  83.21 & 57.74 \\
         Attention & Constant Update & 93 & 72.84 \\
         & Layerwise Update & \textbf{93} & \textbf{73.35} \\
         \hline
         & None & 81.63 & 59.45 \\
         Jacobian & Constant Update & 92.58 & 73.71 \\
         & Layerwise Update & \textbf{93.24} & \textbf{73.84} \\
         \hline
         & None & 84.74 & 56.23 \\
         Hessian & Constant Update & 93 & 66.71\\
         & Layerwise Update& \textbf{93.14} & \textbf{69.98} \\
         \hline
    \end{tabular}
    \caption{Results for CIFAR Experiments. The Constant Update imply the learning rate scheduler used at the 25th and 35th epoch. The layerwise update implies the learning rate optimized per crucial layer at every 25th epoch interval. }
    \label{tab:cifar}
\end{table*}

\begin{figure*}[t]
    \centering
    \subfloat[\centering ]{{\includegraphics[width=0.9\linewidth]{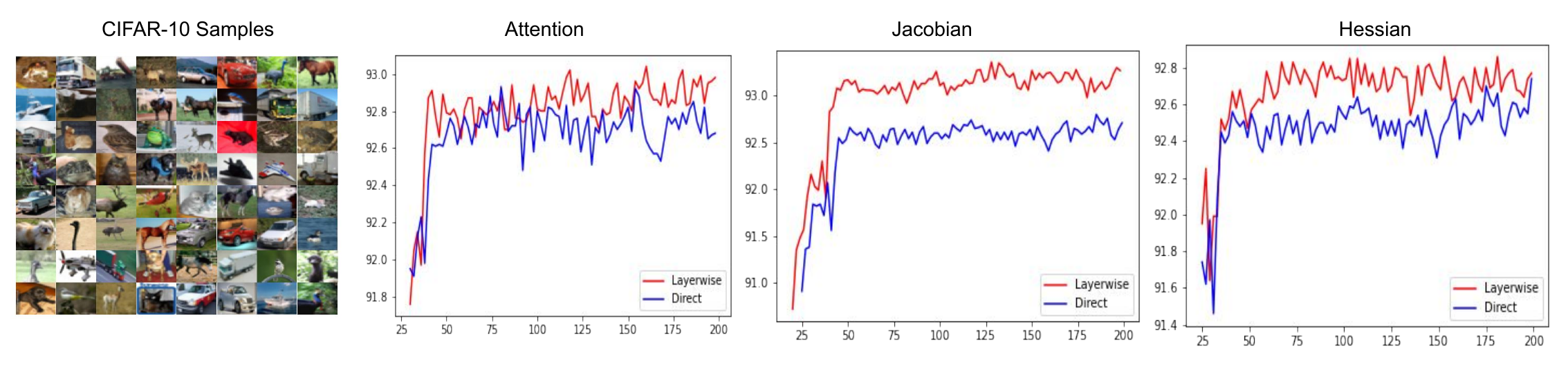} }}
    \quad
    \subfloat[\centering ]{{\includegraphics[width=0.9\linewidth]{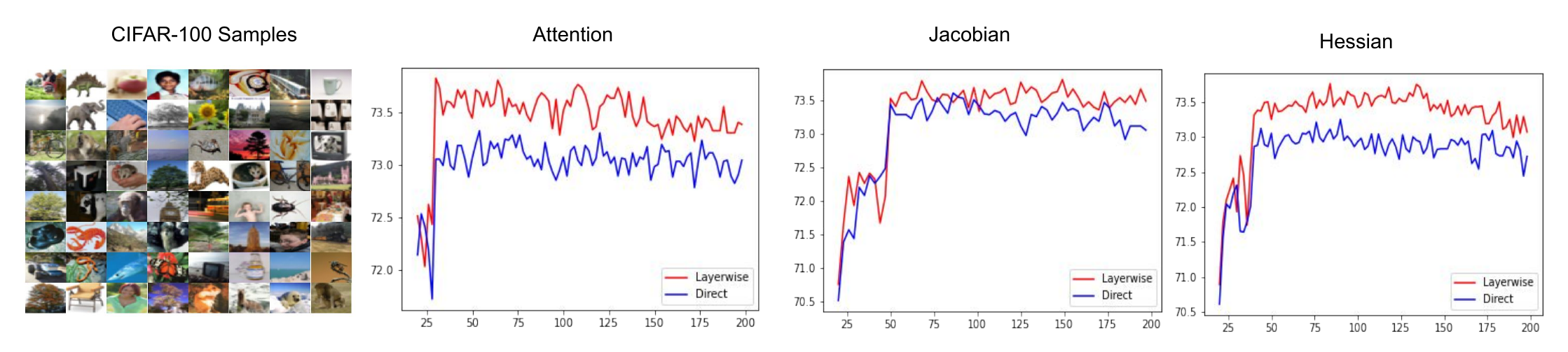}}}
    \quad
    \subfloat[\centering ]{{\includegraphics[width=0.9\linewidth]{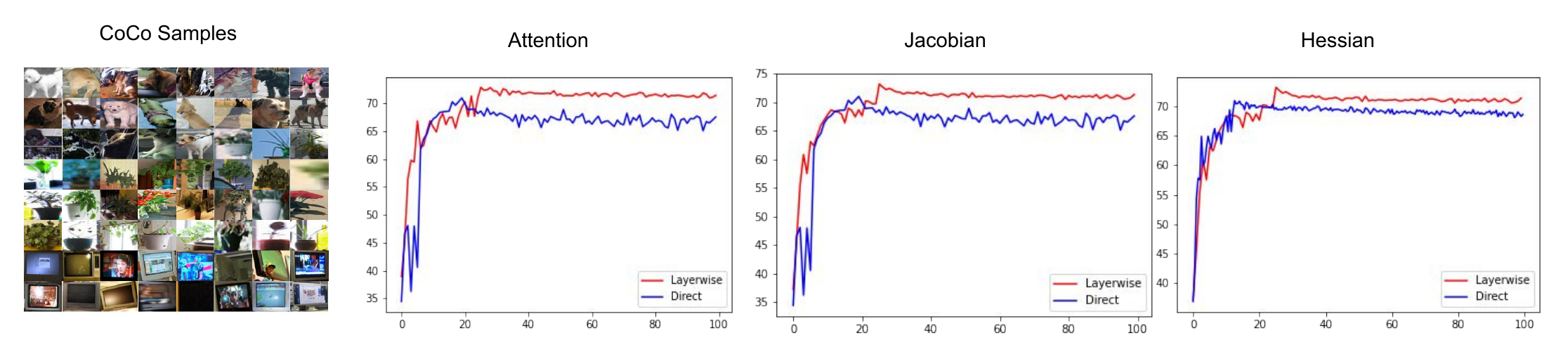}}}
    \caption{Learning curves (epoch vs. accuracy)  for with and without layerwise learning in attention-map, jacobian, and hessian-based knowledge distillation. (a) Learning curves along with some samples from the CIFAR-10 dataset (b) Learning curves along with some samples from the CIFAR-100 dataset (c) Learning curves along with some samples from the CoCo dataset. For CIFAR experiments, `Direct' refers to `Constant Update' in Table 1, whereas in the CoCo experiment, `Direct' refers to None in Table 2}
    \label{fig:results}
\end{figure*}

\subsection{CoCo Experiments}

We performed multi-class classification experiments with over 80 different categories. In the CoCo dataset, we have 562k training images and 36k test images. The images are preprocessed with random cropping and resizing to three channels of 224x224 size.  The models we have assigned for the student-teacher training are ResNet50 as the teacher model and ResNet34 as the student model. In a similar fashion to CIFAR, we calculate Attention, Jacobian, and Hessian. In a similar fashion to CIFAR, we calculate Attention, Jacobian, and Hessian. More effectively, we only calculate the maps at the crucial conversion layers of the ResNet, where we observe a dimensional change in the feature channels. Analogous to the CIFAR Experiments, we have selected 4 layers for the mapping. We train both the student and teacher model for 100 epochs.

We performed experiments across the three transfer learning methods with: 
\begin{itemize}
    \item Constant Learning Rate. 
    \item Layerwise Learning Rate. 
\end{itemize}

\section{Results}

We observed consistent improvement in student model performance while we adapted individual layer-wise learning rates based on our proposed method. We observed such a performance boost for both attention-map and order-based (Jacobian and Hessian) knowledge distillation against all three datasets. Furthermore, we also observed that the contribution of layer-wise learning with our method is more consistent when the learning task becomes harder. For instance, the 100-class classification in CIFAR-100 is harder than the 10-class classification in CIFAR-10. The CoCo multi-class classification is even harder than single-class classifications in CIFAR datasets. We observed the highest and most stable improvement in CoCo multi-class image classification (Figure \ref{fig:results}). In the following sections, we describe the results for each individual dataset in detail.

\begin{table}[!h]
    \centering
    \begin{tabular}{|c|c|c|}
    \hline
        Method &  LR update & CoCo Accuracy\\
        \hline
        Attention & None &  69.9 \\
          & Layerwise & \textbf{72.9}  \\
         \hline
         Jacobian & None & 70.10 \\
         & Layerwise & \textbf{73.14}  \\
         \hline
        Hessian & None & 69.82 \\
         & Layerwise & \textbf{73.08} \\
         \hline
    \end{tabular}
    \caption{Results for CoCo Experiments. The comparison is across no learning rate scheduler and the proposed learning rate optimizer.}
    \label{tab:cifar}
\end{table}

\subsection{CIFAR-10 Experiments}
For single class image classification on the CIFAR-10 dataset consisting of samples from 10 distinct categories, the teacher model had an accuracy of 93.84\% with a learning rate scheduler at constant intervals. Since the maximum accuracy a student model can achieve theoretically is the accuracy of the teacher model, the accuracy of the student model was upper bounded by 93.84\%. The CIFAR-10 dataset has only 10 classes and the classification task is not much difficult for the student models themselves. As a result, we observe close to teacher model accuracy in our attention, Jacobian, and Hessian-based student models in CIFAR-10 experiments. Particularly, for attention map-based learning the student model achieves 93\% accuracy (Table 1) even with a learning rate scheduler at constant intervals. Hence, the layer-wise learning rate update could not make much improvement. However, for Jacobian and Hessian map-based student models, adapting layer-wise learning rates caused a slight improvement. Nonetheless, due to the ease of the task, student models without any layer-wise learning performed close to the teacher model, and the improvement brought by our layer-wise learning could not be more than minimal.  

\subsection{CIFAR-100 Experiments}
For single class image classification on the CIFAR-100 dataset consisting of samples from 100 distinct categories, the maximum accuracy obtained with the teacher was 74.78\% with learning rate updates at constant intervals. Classifying images into 100 distinct categories is much harder than classifying them into 10 classes, and so the CIFAR-100 classification task is much harder than that of CIFAR-10. In such a scenario, the accuracy improvement caused by our layer-wise learning rate adaption scheme over the naive constant learning rate update is more visible and stable  (Figure \ref{fig:results}). The closest accuracy to the teacher model (73.84\%) was achieved with Jacobian map-based knowledge distillation with our layer-wise learning rate-based scheme.

\subsection{CoCo Experiments}
The multi-class image classification is much harder as a learning task compared to single-class classifications. Consequently, the learning task in CoCo experiments can be considered `harder' than in CIFAR experiments. In CoCo experiments, the teacher model had an accuracy of 74.89\% without any constant learning rate update. The constant learning rate update did not bring any improvement in student model accuracy either, so we proceeded with no learning rate update for our student models in CoCo experiments as baselines. The accuracy achieved with the student models was much lower than that of the teacher model compared to CIFAR experiments. However, after adapting the layer-wise learning rate update scheme proposed by us, we observed significant improvement in the student model accuracy. Using layer-wise learning rate update improved the accuracy of the attention-based model by 3\%, Jacobian-based model by 3.14\%, and Hessian-based model by 3.26\%. These are the highest improvements across CIFAR-10, CIFAR-100, and CoCo experiments. This suggests that the contribution of layer-wise learning rate becomes more evident when the learning task becomes harder. From Figure \ref{fig:results}, we also observe that the improvement by layer-wise learning rate is more stable and consistent in the case of CoCo compared to those of CIFAR datasets. In addition, for the CoCo experiment, the closest accuracy to the teacher model (73.14\%) was achieved with the Jacobian map-based knowledge distillation with our layer-wise learning rate-based scheme. This situation is similar to that of CIFAR-100. It might be the case that Jacobian performs better than attention map-based distillation for harder learning tasks. However, more investigation is required in this direction to draw such a conclusion.

\section{Conclusion}\label{sec:Conclusion}
Knowledge distillation from large teacher neural networks to smaller student networks has recently become popular in transfer learning research. The state-of-the-art methods use attention map or parameter derivative (Jacobian or Hessian) maps to perform knowledge distillation from the teacher model to the student model. In this paper, we propose a new scheme for updating the learning rate per critical student model layer with the help of the teacher model. This scheme can be easily added to the knowledge distillation methods. We also demonstrate that adding such a scheme improves the student model accuracy for all the attention-map, Jacobian, and Hessian map-based knowledge distillation methods across diverse benchmark datasets. We also observe that our layer-wise learning schemes become more effective when the learning task becomes harder. Our experimental results suggest that our layer-wise learning method would bring a significant contribution to transfer learning research.

\section{Acknowledgement}\label{sec:funding}
This work was supported in part by U.S. NIH grants R01GM134020 and P41GM103712, NSF grants DBI-1949629, DBI-2238093, IIS-2007595, IIS-2211597, and MCB-2205148. This work was supported in part by Oracle Cloud credits and related resources provided by Oracle for Research, and the computational resources support from AMD HPC Fund. MRU were supported in part by a fellowship from CMU CMLH.

\bibliographystyle{unsrt}
\bibliography{main}
\end{document}